\documentclass[acmsmall]{acmart}

\setcopyright{none}
\copyrightyear{2026}
\acmYear{2026}
\acmDOI{}

\usepackage{xurl}
\usepackage{hyperref} 
\begin{document}

\title{Talking to Extraordinary Objects: Folktales Offer Analogies for Interacting with Technology}
\author{Martha Larson}
\email{martha.larson@ru.nl}
\orcid{1234-5678-9012}
\affiliation{%
  \institution{Radboud University}
  \city{Nijmegen}
  \country{Netherlands}
}

\renewcommand{\shortauthors}{Martha Larson}




\maketitle
The introduction of new technology goes hand in hand with new analogies.
In the 19\textsuperscript{th} century, the steam locomotive was widely referred to as the `iron horse', a comparison with existing transport. 
Charles Dickens used the phrase `mad dragon' in his account of riding an American train in the era.\footnote{\url{https://www.gutenberg.org/ebooks/675}} 
New computing technology adopts analogies to describe of information transmission, storage, and retrieval (`packet', `cloud', `search engine').
Beyond these names, which reflect function and form, are also analogies that guide people in using the technology.
Point and click interaction with personal computers built on the `desktop metaphor', an analogy that prompts users to manipulate files and documents analogously to a physical office.

By the 1990's, technological advances had led to keen anticipation of new interaction paradigms based on speech and language.
The reasoning was appealing.
The most direct way for people to communicate with technology was the way that they communicated with each other---everyone is an expert when it comes to talking~\cite{rudnicky94speechtech}.
This idea quickly morphed into the vision of an interface that would act like a butler, in analogy with an actual human~\cite{vandam97postwimp}.
The butler opened the door, so to say, for analogies to come.
The next generation of analogies for interacting with technology using speech and language promises to be more extraordinary, more diverse, and possibly much less human.

\section{Anthropomorphization Meets its Moment of Reckoning}
Today's information technology includes applications for search, recommendation, decision support, communication support, education, and companionship.
Robotics connects these functions with sensorimotoric capacity. 
These applications are extraordinary in that they go beyond what humans can do.

We find ourselves at a moment of reckoning with anthropomorphized interaction with technology.
The analogy with human-to-human interaction not longer fits well when we value technology for its ability to exceed humans in speed, stamina, perception, pattern matching ability, and scope of the information it can process.
Evidence of the downsides, disadvangtages, and even dangers of anthropomorphism are accumulating~\cite{akbulut24toohuman}.

De facto standards for interfaces are known to support new users, ease use, and allow skills to transfer between applications~\cite{vandam97postwimp}.
The moment of reckoning for anthropomorphization is leading to a turn away from a single universal analogy for interacting with technology, a rejection of computer-as-human as the de facto standard.
However, if technology should not mimic humans, what else is there?

The answer provided by the concept of `otherware' is that technology should emphasize its otherness, i.e., the way in which it is not human, rather than aiming to appear human~\cite{hassenzahl20otherware}.
Robotics has proposed to treat robots as animals~\cite{darling22newbreed}.
Ideal would be analogies that maintain language, intelligence, and extraordinary abilities, but eliminate other aspects of human-to-human interaction, such as social obligation and reciprocity.

In fact, we are all heirs to a rich domain in which extraordinary entities abound and language and intelligence can decouple from humanness.
The world of folktales offers exactly such analogies.
Folktales are stories originating in oral tradition that are passed down over generations.
Some have been estimated to be thousands of years old~\cite{dasilva16phylogenetic}.
As such, they are representative of the collective imagination of humankind.

The world of folktales is populated with extraordinary entities whose abilities go far beyond what humans can achieve in the real world.
Some entities are human-like or animal-like.
Other entities are clearly objects.
We are as free as Dickens to turn to folklore for analogy.
His dragon is a fantastic beast familiar from folklore around the world---the effectiveness of the analogy is not impeded by the fact that dragons are imaginary.

Not all extraordinary entities speak, but many do. 
In particular, there is nothing remarkable about talking to extraordinary objects, or these objects displaying intelligence.
The promise of analogies for information technology motivates an unexpected journey into a world of folktales in search of inspiration and insight. 

\section{The Relevance of Folktales for Interaction with Technology}
Folktales engage with a wide spectrum of themes, including courage, love, integrity, responsibility.
The tales also touch on anthropomorphization. 
In a classic folktale, `Little Red Riding Hood', a non-human entity being taken as human is central to the story.
Across the versions of the story, a constant is the evil wolf presenting himself to Red Riding Hood as her grandmother.
Here, we follow the Grimm version.\footnote{\protect\url{https://www.gutenberg.org/cache/epub/5314/pg5314-images.html\#chap26}}
The wolf is lying in the grandmother's bed, and Red Riding Hood and the wolf engage in an exchange,
\begin{quote}
``Oh! grandmother,” she said, ``what big ears you have!''

``The better to hear you with, my child,'' was the reply. 
\end{quote}

She asks about the eyes and hands, and finally, when she asks about the mouth, she realizes the wolf is not her grandmother.
It is too late, and the wolf jumps out of bed and swallows her.
Calls for technology to be limited in its ability to imitate humans could have been inspired by this tale.

Folktales also engage with extraordinary, beyond-human ability to access information. 
The danger of misuse of this ability is a theme of `Little Snow-white'. 
Again, we follow the Grimm version\footnote{\protect\url{https://www.gutenberg.org/cache/epub/5314/pg5314-images.html\#chap53}}
The jealous queen of the story receives daily updates from a mirror. 
Every day she asks,
\begin{quote}
    ``Looking-glass, Looking-glass, on the wall,
    
Who in this land is the fairest of all?''
\end{quote}

and the mirror replies, assuring that queen that she is most beautiful.
The queen becomes murderous the day the mirror changes its answer and names Snow White instead. 
Calls to discourage technology that reinforces harmful or unethical behavior could have been inspired by this tale.

\section{Folktales as a Sandbox of the Imagination}
Folktales have been collected by campaigns across the centuries. 
Widely known is the work in Germany of the Grimm Brothers in the 19\textsuperscript{th} century~\cite{georgeandjones95folkloristics}.
Research has established that folktales occur of similar stories, which can cross languages, regions, and even cultures.
Folktales are born as oral stories, but can continue to evolve in interaction with the written medium as well as with theater, film and other multimedia.

Folktales are often discussed on the basis of specific texts, as  we do here, but the tales exist, in their essence, in abstract form beyond individual retellings.
Further, the existing texts do not reflect precious tales that might have been lost over time.
As a result, the usefulness of folktales for information technology is unlikely to arise from statistical analysis, but instead through inspiration on the basis of key examples.

Folktales often start with the phrase, `Once upon a time', which signals an existence beyond place and time. 
In this way, they contrast with other orally transmitted stories, including fables, which communicate a moral, and legends, which purport to be history.
Folktales do not reduce to education, entertainment, or calcified tradition. 

Recently, folktales have been interpreted as a sort of experimental space for development of ethical understanding~\cite{mickinell19ethics}.
As such, they are a source of inspiration, but we must resist turning to them for absolute truths.
Their bursts of sexism and racism as well as their potential for abuse to promote unethical messages~\cite{mickinell19ethics} makes them unsuited as a source of specific guidance.
Instead, their value lies in their longstanding role as a sandbox-type space in which the human imagination can test ideas through innovation and manipulation.
They should not be dismissed as irrelevant or discounted as ``low culture''.

``Their tales still read like innovative strategies for survival,'' writes folklorist Jack Zipes in the introduction to his translation of the tales of the Grimm Brothers~\cite{zipes03grimm}.
People turn to folktales in cases of need.
Like our ancestors, we enter the world of folktales because we are in need of inspiration for new analogies of for interacting with technology.

\section{Extraordinary Objects Communicate with Natural Language}
In the world of folktales, the use of natural language is everywhere.
In the English tale `Jack and the Beanstalk',\footnote{\protect\url{https://www.gutenberg.org/cache/epub/7439/pg7439-images.html\#link2H_4_0015}} the hero Jack is hiding from an ogre who shouts:
\begin{quote}
``Fee-fi-fo-fum, I smell the blood of an Englishman,

Be he alive, or be he dead, I'll have his bones to grind my bread.''
\end{quote}
The ogre has human form, so language could be considered a natural extension to him being a human-like extraordinary entities.

Animals talk in folk tales.
In the African American tradition, the trickster hero rabbit engages verbally with the devious fox who has caught him:
\begin{quote}
``Drown me! Roast me! Hang me! Do whatever you please,'' said Brer Rabbit. ``Only please, Brer Fox, please don’t throw me into the briar patch.''\footnote{\protect\url{https://www.americanfolklore.net/brer-rabbit-and-the-tar-baby}}
\end{quote}
Of course, briars are actually a rabbit habitat, and the fox is foiled.
Talking animals can be evil like Red Riding Hood's wolf or good as in the `The Story of the Three Bears',\footnote{\protect\url{https://www.gutenberg.org/cache/epub/7439/pg7439-images.html\#link2H_4_0020}} a tale in which some, but not all, versions involve a girl named Goldilocks.

Extraordinary objects also have language capabilities.
The ogre in `Jack and the Beanstalk' has a golden harp that plays when he commands `Sing!'
When Jack tries to steal the harp, it calls out to the ogre `Master! Master!'

Extraordinary beings, which are human-like or animals, and extraordinary objects are sharply distinct, although both may have language capabilities.
Extraordinary objects are possessions, e.g., the harp belongs to the ogre, and can be picked up and carried away.
Extraordinary objects are highly valuable, but their value is associated with their superhuman ability to do a task---they are supremely useful.
They do not have intrinsic moral worth, in contrast to a human or a pet animal.
The story makes clear the harp is being stolen by Jack, not kidnapped.
 
A simple parameterization allows the organization of extraordinary objects in terms of language capability.
Some do not use language, like the mirror from certain versions of `Beauty and the Beast', a French tale,\footnote{\protect\url{https://www.gutenberg.org/cache/epub/7074/pg7074-images.html}} 
which shows the heroine her father. 

Of those extraordinary objects that use language, some only respond to language and do not themselves speak. 
The cave in `Ali Baba and the Forty Thieves'\footnote{\protect\url{https://www.gutenberg.org/ebooks/61974}}, from the `Book of the Thousand and One Nights' opens in a response to the fixed command, `Open sesame'.
The Norse folktale, `Why the Sea is Salt'\footnote{\protect\protect\url{https://www.gutenberg.org/cache/epub/8933/pg8933-images.html\#chap02}} is the story of a hand mill that is capable of producing anything it is asked for.
The mill responds to free-form requests, rather than fixed commands.
Critically, it is important to know how to talk to it.
When it lands in the hands of a skipper without this knowledge, it grinds so much salt that it sinks his ship and turns the whole ocean to saltwater. 

Other objects both respond to language and produce it themselves.
The orgre's harp responds to the command, `Sing!', but also can call out.
Among such examples are objects that engage in conversation. 
In the Norse folktale `Boots and his Brothers',\footnote{\protect\url{https://www.gutenberg.org/cache/epub/8933/pg8933-images.html\#chap46}} the hero, Askeladden (translated here as `Jack') greets an axe in the woods:
\begin{quote}
“Good day!” said Jack. “So you stand here all alone and hew, do you?”

“Yes; here I’ve stood and hewed and hacked a long long time, waiting for you”, said the Axe.

“Well, here I am at last”, said Jack, as he took the axe, pulled it off its haft, and stuffed both head and haft into his wallet. 
\end{quote}

In short, in folktales language does not necessarily imply human-like-ness.
Both extraordinary beings and extraordinary objects talk. 
They provide inspiration for analogies with non-human others--- `otherware'~\cite{hassenzahl20otherware} requires such analogies.
They also offer insight.
No one specific language style is used when talking to extraordinary objects, reflecting the natural capacity of humans for adaptability in spoken communication~\cite{cooke14talker}.
Askeladden addresses the axe with `Good day!' but, looking across tales, the need for politeness or conversational style is not a given.
And indeed, once Askeladden has taken the axe into his possession, he addresses it with commands.

\section{Extraordinary Objects Display Intelligence}
The world of folktales offers examples of objects that display intelligence, and again a simple parameterization supports organization. 
First, there are extraordinary objects that are intelligent in that they can recognize ownership or intent, like the ogre's harp. 
In the tale, `The Ass, the Table, and the Stick',\footnote{\protect\url{https://www.gutenberg.org/cache/epub/7439/pg7439-images.html\#link2H_4_0041}}
the stick responds to the command, `Up stick and bang him!' by attacking whatever enemy of the stick's owner is referring to.
It recognizes ownership and intent and does not attack the owner himself nor mistake the designated foe.

In contrast, some extraordinary objects do not distinguish individuals and cannot interpret intent.
The cave in `Ali Baba' will open to anyone who says, `Open sesame'. 
In the Norse tale, `Soria Moria Castle',\footnote{\protect\url{https://www.gutenberg.org/cache/epub/8933/pg8933-images.html\#chap56}} the hero has a wishing ring.
He makes a rhetorical statement in the form of a wish, and the ring fulfills the wish although he did not intend it to.

Finally, there are extraordinary objects that are intelligent in that they are sensitive to ethical use.
In the West African folktale `The Grinding Stone that Ground Flour by Itself',\footnote{\protect\url{https://www.gutenberg.org/cache/epub/66923/pg66923-images.html\#ch13}} a grinding stone provides a supply of food and drink. The stone warns the greedy spider against stealing it,
\begin{quote}
``Spider, spider, put me down,'' said the stone.

``The pig came and drank and went away,

The antelope came and fed and went away:

Spider, spider, put me down.''
\end{quote}
In contrast, the mirror of `Snow White', does not object to repeatedly answering the question.
Instead, it feeds the queen's jealousy and directs her murderous attention to Snow White. 

In short, intelligence is manifested by objects in folktales, offering inspiration for analogies.
We also gain insight on how language capability can decouple from  intelligence. 
The stick that responds to the fixed-command `Up stick and bang him!' has limited language capability compared to the wishing ring, which responds to the full range of natural language. 
Yet the stick is more intelligent in that it can interpret intent. 

\section{Be Careful What You Wish For}
Our journey into the world of folktales has uncovered potential analogies for interacting with technology---examples of talking to extraordinary objects.
Folktales contribute to the current reckoning with anthropomorphization.
They reinforce our aspiration to develop technology that retains the usefulness and comfort of speech and language interaction but avoids the dangers of relying on a strong association with humanness.

Folktales involve a large number of varied extraordinary objects.
However, it is not particularly difficult to become familiar with what they all do and how to communicate with them---children do it naturally from their first exposure to the tales.
This capacity suggests that we should expect that ultimately people will be cognitively conformable with an entire continuum of otherness~\cite{hassenzahl20otherware}, rather than striving for a single de facto standard of highly human-like interaction with technology.

A journey into the world of folktales could conceivably transform the dream of a butler into the dream of an extraordinary lamp.
In `Alaeddin; or, The Wonderful Lamp',\footnote{\protect\url{https://www.gutenberg.org/cache/epub/60889/pg60889-images.html\#Page_49}} an oil lamp fulfills wishes expressed in spoken form.
Digging deeper in the tale, the extraordinary ability of Aladdin's lamp is due to a human-like genie, who is `slave of the lamp.'
We must never let analogies with extraordinary objects conceal the amount of human work that is behind technology, and the nature of the contributions of the people who make it possible for technology to talk.

Like many folktales, `Why the Sea is Salt' treats  themes of moderation and of being careful for what you wish for.
The extraordinary mill is an object.
But its lack of human-like motivations did not prevent it from altering the ocean across the entire planet.
Successful anthropomorphic reckoning does not absolve us of the need to consider the global consequences of the technology we develop.
Also, moderation must remain easy and natural.
We need the option to put technology in a corner, like an object, and only bring it out when we absolutely need it.

\bibliographystyle{ACM-Reference-Format}
\bibliography{bibfile}

@book{georgeandjones95folkloristics,
  title     = "Folkloristics: An Introduction",
  author    = "Georges, Robert A. and Jones, Michael Owen",
  year      = 2003,
  publisher = "Indiana University Press",
  address   = "Bloomington, Indiana"
}

@book{zipes03grimm,
  title     = "The Complete Fairy Tales of the Brothers Grimm",
  author    = "Zipes (translator), Jack",
  year      = 2003,
  publisher = "Bantam Books",
  address   = "New York, New York"
}

@book{darling22newbreed,
  title     = "New Breed: What Our History With Animals Reveals About our Future With Robots",
  author    = "Darling, Kate",
  year      = 2022,
  publisher = "Holt Paperbacks",
  address   = "New York, New York"
}

@article{akbulut24toohuman, title={All Too Human? Mapping and Mitigating the Risk from Anthropomorphic AI}, volume={7},  abstractNote={The development of highly-capable conversational agents, underwritten by large language models, has the potential to shape user interaction with this technology in profound ways, particularly when the technology is anthropomorphic, or appears human-like. Although the effects of anthropomorphic AI are often benign, anthropomorphic design features also create new kinds of risk. For example, users may form emotional connections to human-like AI, creating the risk of infringing on user privacy and autonomy through over-reliance. To better understand the possible pitfalls of anthropomorphic AI systems, we make two contributions: first, we explore anthropomorphic features that have been embedded in interactive systems in the past, and leverage this precedent to highlight the current implications of anthropomorphic design. Second, we propose research directions for informing the ethical design of anthropomorphic AI. In advancing the responsible development of AI, we promote approaches to the ethical foresight, evaluation, and mitigation of harms arising from user interactions with anthropomorphic AI.}, number={1}, journal={Proceedings of the AAAI/ACM Conference on AI, Ethics, and Society}, author={Akbulut, Canfer and Weidinger, Laura and Manzini, Arianna and Gabriel, Iason and Rieser, Verena}, year={2024}, month={Oct.}, pages={13-26} }

@article{cooke14talker,
title = {The listening talker: A review of human and algorithmic context-induced modifications of speech},
journal = {Computer Speech and Language},
volume = {28},
number = {2},
pages = {543-571},
year = {2014},
issn = {0885-2308},
author = {Martin Cooke and Simon King and Maëva Garnier and Vincent Aubanel}
}

@article{mickinell19ethics,
author={McKinnell,Liz},
year={2019},
month={04},
title={The Ethics of Enchantment: The Role of Folk Tales and Fairy Tales in the Ethical Imagination},
journal={Philosophy and Literature},
volume={43},
number={1},
pages={192-209},
isbn={01900013},
language={English},
}

@article{hassenzahl20otherware,
author = {Hassenzahl, Marc and Borchers, Jan and Boll, Susanne and P\"{u}tten, Astrid Rosenthal-von der and Wulf, Volker},
title = {Otherware: how to best interact with autonomous systems},
year = {2020},
issue_date = {January - February 2021},
publisher = {Association for Computing Machinery},
address = {New York, NY, USA},
volume = {28},
number = {1},
issn = {1072-5520},
journal = {Interactions},
month = dec,
pages = {54–57},
numpages = {4}
}

@article{vandam97postwimp,
author = {van Dam, Andries},
title = {Post-WIMP user interfaces},
year = {1997},
issue_date = {Feb. 1997},
publisher = {Association for Computing Machinery},
address = {New York, NY, USA},
volume = {40},
number = {2},
issn = {0001-0782},
journal = {Communications of the ACM},
month = feb,
pages = {63–67},
numpages = {5}
}

@article{rudnicky94speechtech,
author = {Rudnicky, Alexander I. and Hauptmann, Alexander G. and Lee, Kai-Fu},
title = {Survey of current speech technology},
year = {1994},
issue_date = {March 1994},
publisher = {Association for Computing Machinery},
address = {New York, NY, USA},
volume = {37},
number = {3},
issn = {0001-0782},
journal = {Communications of the ACM},
month = mar,
pages = {52–57},
numpages = {6}
}

@article{dasilva16phylogenetic,
    author = {da Silva, Sara Graça and Tehrani, Jamshid J.},
    title = {Comparative phylogenetic analyses uncover the ancient roots of Indo-European folktales},
    journal = {Royal Society Open Science},
    volume = {3},
    number = {1},
    pages = {150645},
    year = {2016},
    month = {01},
    issn = {2054-5703}
}

\end{document}